\documentclass[letterpaper, 10 pt, conference]{ieeeconf}

\makeatletter
\def\endthebibliography{%
  \def\@noitemerr{\@latex@warning{Empty `thebibliography' environment}}%
  \endlist
}
\makeatother
 
\IEEEoverridecommandlockouts
\usepackage{cite}
\usepackage{amsmath,amssymb,amsfonts}
\usepackage{graphicx}
\usepackage{textcomp}
\usepackage{xcolor}
\usepackage{todonotes}
\usepackage{amsmath}
\usepackage{leftidx}
\usepackage{siunitx}
\usepackage{multirow}
\usepackage{footnote}
\usepackage[utf8]{inputenc}
\usepackage{array}
\usepackage{mdwmath}
\usepackage{mdwtab}
\usepackage{eqparbox}
\usepackage{eufrak}
\usepackage{url}
\usepackage{cite}
\usepackage{amsmath,amssymb,amsfonts}
\usepackage{algorithm2e}
\RestyleAlgo{ruled}
\usepackage{textcomp}
\usepackage{lettrine}
\usepackage{caption}
\usepackage{subcaption}
\usepackage[export]{adjustbox}
\usepackage{amsmath}
\usepackage{leftidx}
\usepackage{siunitx}
\usepackage{multirow}
\usepackage{balance}
\usepackage[export]{adjustbox}
\usepackage{booktabs}
\usepackage{hyperref}
\usepackage{balance}

\def\BibTeX{{\rm B\kern-.05em{\sc i\kern-.025em b}\kern-.08em
    T\kern-.1667em\lower.7ex\hbox{E}\kern-.125emX}}
\begin{document}

\title{\LARGE \bf Graph-based Global Robot Simultaneous Localization and Mapping using Architectural Plans}

\author{Muhammad Shaheer$^{1}$, Jose Andres Millan-Romera$^{1}$, Hriday Bavle$^{1}$, Jose Luis Sanchez-Lopez$^{1}$,\\ Javier Civera$^{2}$ and Holger Voos$^{1}$ 
\thanks{$^{1}$Authors are with the Automation and Robotics Research Group, Interdisciplinary Centre for Security, Reliability and Trust (SnT), University of Luxembourg (UL). Holger Voos is also associated with the Faculty of Science, Technology and Medicine, University of Luxembourg, Luxembourg.
\tt{\small{\{muhammad.shaheer, jose.millan, hriday.bavle, joseluis.sanchezlopez, holger.voos\}}@uni.lu}}%
\thanks{$^{2}$Author is with I3A, Universidad de Zaragoza, Spain
{\tt\small jcivera@unizar.es}}%
 \thanks{*This work was partially funded by the Fonds National de la Recherche of Luxembourg (FNR) under the project 17097684/RoboSAUR, by a partnership between the SnT-UL and Stugalux Construction S.A., and by the Spanish and Aragón governments (projects PID2021-127685NB-I00, TED2021-131150B-I00 and DGA\_FSE-T45\_20R).
 For the purpose of Open Access, the author has applied a CC BY 4.0 public copyright license to any Author Accepted Manuscript version arising from this submission.
 }
}

\maketitle
\begin{abstract}

\label{abstract}
In this paper, we propose a solution for graph-based global robot simultaneous localization and mapping (SLAM) using architectural plans. 
Before the start of the robot operation, the previously available architectural plan of the building is converted into our proposed architectural graph (A-Graph).
When the robot starts its operation, it uses its onboard LIDAR and odometry to carry out an online SLAM relying on our situational graph (S-Graph), which includes both, a representation of the environment with multiple levels of abstractions, such as walls or rooms and their relationships, as well as the robot poses with their associated keyframes.
Our novel graph-to-graph matching method is used to relate the aforementioned S-Graph and A-Graph, which are aligned and merged, resulting in our novel informed Situational Graph (iS-Graph).
Our iS-Graph not only provides graph-based global robot localization, but it extends the graph-based SLAM capabilities of the S-Graph by incorporating into it the prior knowledge of the environment existing in the architectural plan.


\end{abstract}

\section{Introduction}
\label{introduction}


The construction industry is increasingly using mobile robots, which offer numerous potential benefits. These robots can significantly reduce costs by regularly inspecting ongoing construction sites and monitoring progress. However, most robots used in construction are either teleoperated or operate semi-autonomously, primarily due to the perception challenges associated with the ever-changing nature of construction sites. To enable fully autonomous operation, it would be advantageous for these robots to possess comprehensive prior knowledge of the construction site's geometry. By combining this prior knowledge with sensor readings during real-time operation, these robots could achieve robust and accurate global localization within construction sites.

Digital architectural plans, such as Building Information Modelling (BIM) \cite{bim}, provide a means of capturing and  communicating information about a construction site and incorporating it as prior knowledge about the scene. Works such as \cite{connecting_semantic_bim, bim_localization, robot_localization_shaheer} have addressed the problem of extracting relevant structural knowledge from BIM and using it for real-time robot localization. However, these methods only extract geometric information from the BIM and do not leverage the topological and relational information also available in it, which limits the robustness and accuracy in complex and changing construction sites. 

To tackle this problem, we present a novel approach to localize robots leveraging not only geometry but also higher-level hierarchical information from architectural plans. We present in this paper how to model the BIM information in the form of a graph that we denote as Architectural Graph (\textit{A-Graph}), and then match and merge with the online Situational Graph (\textit{S-Graph}) \cite{s_graphs}, \cite{s_graphs+} that the robot builds as it navigates the environment. As a key aspect, translating low-level geometry into high-level features in both graphs is what allows a robust matching between such different inputs.

Our method can be divided into three main stages. In the first one, an \textit{A-Graph} is created for a given environment with nodes representing the semantic features available in a BIM model, specifically, wall-surfaces, doors, and rooms as the graph nodes, and edges containing the relevant relational information such as two wall surfaces comprising a wall, four wall surfaces connecting room and rooms connected through doors. In the second stage, running in real-time onboard the robot, a \textit{S-Graph} is estimated using 3D LiDAR measurements. The nodes of our \textit{S-Graph} correspond to semantics such as wall surfaces and rooms, and the edges correspond to constraints between these wall surfaces and the relevant room nodes. Finally, to localize the robot within this environment, a graph-matching algorithm is proposed utilizing hierarchical information from both graphs to provide the best match candidates finally resulting in informed (\textit{iS-Graphs}) that fuses the information of both. This last graph will be the one used for global localization.


\begin{figure*}[t]
    \centering
    \includegraphics[width=0.8\textwidth]{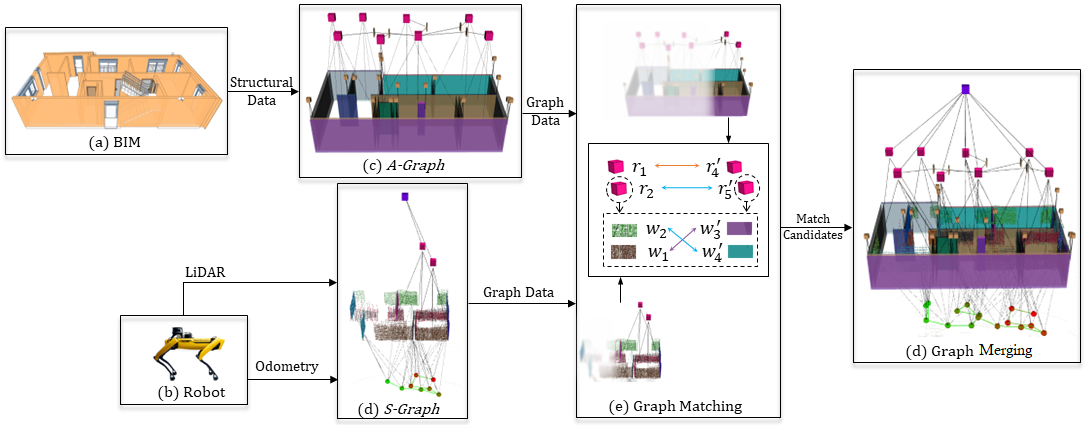}
\caption{Overview of our approach. We generate offline an Architectural Graph (\textit{A-Graph}) from a BIM model. A robot estimates online a Situational Graph (\textit{S-Graph}) from its sensors. We do graph matching between the two, align them, and merge their information. This generates the final \textit{iS-Graphs}, which is utilized by the robot to be localized with respect to the BIM.}
 \label{fig:intro_pic}
\end{figure*}

\section{System Overview}
\label{system_overview}
Fig.~\ref{fig:intro_pic} shows an overview of the proposed approach. Firstly the BIM information is extracted for a given environment to create the two-layered \textbf{Architectural Graphs} (Section.~\ref{sec:a_graphs}) in an offline manner, with all the elements extracted in the BIM frame of reference $B$. As the robot navigates within the given environment, an online \textbf{Situational Graph} (Section.~\ref{subsec:situational_graphs}) is estimated by the robot in the map frame of reference $M$. In parallel, we run our the \textbf{Graph Matching} method (Section.~\ref{sec:graph_matching}) to provide match candidates between the existing \textit{A-Graphs} and the current \textit{S-Graph}. Finally, after retrieving the best match candidate, our \textbf{Graph Merging} (Section.~\ref{sec:graph_merging}) provides the merged \textit{iS-Graph} utilized for the global localization with respect to the BIM frame of reference $B$.


\subsection{Situational Graphs (S-Graphs)} \label{subsec:situational_graphs}

\textit{S-Graphs} are four-layered optimizable hierarchical graphs built online using 3D LiDAR measurements. The full details of the \textit{S-Graphs} we use in this work can be found in \cite{s_graphs},\cite{s_graphs+}. In brief, their four layers can be summarized as:

\textbf{Keyframes Layer.} It consists of the robot poses factored as $\leftidx{^M}{\mathbf{x}}_{R_i} \in SE(3)$ nodes in the map frame $M$ with pairwise odometry measurements constraining them. 

\textbf{Walls Layer.} It consists of the planar wall-surfaces $\leftidx{^M}{\boldsymbol{\pi}}_{i} \in \mathbb{R}^3$ extracted from the 3D LiDAR measurements and factored using minimal plane parameterization. The planes observed by their respective keyframes are factored using pose-plane constraints. 

\textbf{Rooms Layer:} It consists of two-wall rooms $\leftidx{^M}{\boldsymbol{\gamma}}_{i} \in \mathbb{R}^2$ or four-wall rooms $\leftidx{^M}{\boldsymbol{\rho}}_{i} \in \mathbb{R}^2$, each constraining either two or four detected wall-surfaces respectively. 

\textbf{Floors Layer:} It consists of a floor node $\leftidx{^M}{\boldsymbol{\xi}}_{i} \in \mathbb{R}^2$ positioned at the center of the current floor level and constraining all the rooms present at that floor level.


\section{Architectural Graphs (A-Graphs)} \label{sec:a_graphs}

We extract relevant information from BIM models into two-layered optimizable graphs denoted as \textit{A-Graphs}. In the lowest-level layer, we will model the geometry of the walls, and in the highest level the rooms. Room-to-wall constraints connect the two layers and neighboring rooms are constrained by doorways. The specific formulation is detailed below.

\textbf{A-Walls Layer.} \label{sec:pose_graph_gen} This layer extracts all the information about the walls and wall surfaces from the BIM and connects them with appropriate wall-to-wall-surface edges. 

\textbf{\textit{Wall-Surfaces:}} Wall-surfaces are planar entities $\leftidx{^{B}}{\boldsymbol{\pi}}$ extracted in the BIM frame of reference $B$. All the wall-surfaces are converted to their Closest Point (CP) representation, as in \cite{s_graphs+}. Wall-surface normals with their component $\leftidx{^B}{{n}_x}$ greater than $\leftidx{^B}{{n}_y}$ are classified as $x$-wall-surfaces, and wall surfaces whose normal component $\leftidx{^B}{{n}_y}$ is greater than the normal component   $\leftidx{^B}{{n}_x}$ are defined as $y$-wall-surfaces. 
These are initialized in the graph as $\leftidx{^B}{\boldsymbol{\pi}} = [\leftidx{^B}\phi, \leftidx{^B}\theta, \leftidx{^B}d]$, where $\leftidx{^B}\phi$ and $\leftidx{^B}\theta$ stand for the azimuth and elevation of the plane in frame $B$ and $\leftidx{^B}d$ is the perpendicular distance in $B$. 

\textbf{\textit{Walls:}} 
We introduce a novel semantic entity with respect to \cite{s_graphs+} in the form of a \textit{Wall} $\boldsymbol{\omega} \in \mathbb{R}^3$, consisting of two planar wall-surfaces. Two opposed planar wall-surface entities either in $x$-direction or $y$-direction with similar perpendicular distance $\leftidx{^B}d$ can be classified as a part of a single wall entity. The wall center $\leftidx{^B}{\boldsymbol{\omega}_{x_i}}$ for two opposed $x$-direction wall-surfaces is computed as:

\begin{gather}
\resizebox{1.\hsize}{!}{$\leftidx{^B}{\mathbf{w}_{x_i}} =  \frac{1}{2} 
 \big[ \lvert {\leftidx{^B}{d_{x_{1}}} \rvert} \cdot \leftidx{^B}{\mathbf{n}_{x_{1}}} - {\lvert \leftidx{^B}{d_{x_{1}}} \rvert} \cdot \leftidx{^B}{\mathbf{n}_{x_{2}}} \big]  + \lvert \leftidx{^B}{{d_{x_{2}}} \rvert} \cdot \leftidx{^B}{\mathbf{n}_{x_{2}}} \nonumber$} \\
 \leftidx{^B}{\boldsymbol{\omega}_{x_i}} = \leftidx{^B}{\mathbf{w}_{x_i}} +  \big[  \leftidx{^B}{\mathbf{\mathfrak{s}}_i} - [\ \leftidx{^B}{\mathbf{\mathfrak{s}}_i} \cdot \leftidx{^B}{\hat{\mathbf{w}}_{x_i}} ] \ \cdot \hat{\leftidx{^B}{\mathbf{w}}_{x_i}} \big]
\end{gather} \label{eq:wall_center}

\noindent where $\leftidx{^B}{\mathbf{\mathfrak{s}}_i} \in \mathbb{R}^3$ is the starting point for a given BIM wall and $\mathbf{n}$ and $d$ are the plane normals and distance. For Eq.~\ref{eq:wall_center} to hold true, all plane normals are converted to point away from the BIM frame of reference as in \cite{s_graphs+}. The wall center along with it wall-surfaces is factored in the graph as:

\begin{multline} \label{eq:infinite_room_node}
    c_{\boldsymbol{\omega}}(\leftidx{^B}{\boldsymbol{\omega}_i},\big[\leftidx{^B}{\boldsymbol{\pi}_{x_{1}}}, \leftidx{^B}{\boldsymbol{\pi}_{x_{1}}}, \leftidx{^B}{\mathfrak{s}}_i]) \\ = \sum_{i=1}^{K} \| \leftidx{^B}{\hat{\boldsymbol{\omega}}_i} - f(\leftidx{^B}{\tilde{\boldsymbol{\pi}}_{x_{1}}}, \leftidx{^B}{\tilde{\boldsymbol{\pi}}_{x_{1}}}, \leftidx{^B}{\mathfrak{s}_i}) \| ^2_{\mathbf{\Lambda}_{\boldsymbol{\tilde{\boldsymbol{\omega}}}_{i,t}}}
\end{multline}

Where $f(\leftidx{^B}{\tilde{\boldsymbol{\pi}}_{x_{1}}}, \leftidx{^B}{\tilde{\boldsymbol{\pi}}_{x_{1}}}, \leftidx{^B}{\mathfrak{s}_i})$ is the function mapping the wall center using the wall-surfaces and its starting point following Eq.~\ref{eq:wall_center}. Wall factors add an additional layer of structural consistency to the graph. A Wall center for opposed planes in $y$-direction is computed following Eq.~\ref{eq:wall_center}.

\textbf{A-Rooms Layer.} The second layer of the graph extracts all the information about the rooms along with the door-ways interconnecting the rooms.   

\textit{\textbf{Rooms:}} We use the similar concept of a four-wall room $\leftidx{^B}{\boldsymbol{\rho}} \in \mathbb{R}^2$ as presented in \cite{s_graphs+}, where each room comprises the four-wall surfaces extracted in the first layer of the graph. 

\textit{\textbf{Door-Ways:}} We incorporate an additional entity in the graph called door-ways interconnecting the room nodes, easily available from BIM. The position of a door-way node $\leftidx{^B}{\boldsymbol{\mathcal{{D}}}} \in \mathbb{R}^3$ is directly extracted from BIM in the frame of reference $B$. Using the semantic information from BIM of the rooms connected by a given door-way, the door-way-to-rooms factor can be formulated as:
\begin{multline}
c_{\boldsymbol{\mathcal{D}}} (\leftidx{^B}{\boldsymbol{\rho}_1},\leftidx{^B}{\boldsymbol{\rho}_2} , \leftidx{^B}{\boldsymbol{\mathcal{D}}_{i}}) = \\ \| \ f(\leftidx{^{B}}{\hat{\boldsymbol{\rho}}_{1}},
\leftidx{^{\rho_1}}{\hat{\boldsymbol{{\mathcal{D}}}}_i})
- \ f(\leftidx{^B}{\hat{\boldsymbol{\rho}}_2}, \leftidx{^{\rho_2}}{\hat{\boldsymbol{\mathcal{D}}}_i}) \|
\end{multline}

Where $\leftidx{^B}{\boldsymbol{\rho}_1}$ and $\leftidx{^B}{\boldsymbol{\rho}_2}$ are the four-wall rooms connected to the door-way $\leftidx{^B}{\boldsymbol{\mathcal{D}}_{i}}$. $\leftidx{^{\rho_1}}{\boldsymbol{\mathcal{D}}_i}$ and $\leftidx{^{\rho_2}}{\boldsymbol{\mathcal{D}}_i}$ are the positions the door-way nodes estimated with respect to rooms $\leftidx{^B}{\boldsymbol{\rho}_1}$ and $\leftidx{^B}{\boldsymbol{\rho}_2}$.

\section{Graph Matching}
\label{sec:graph_matching}

\begin{figure*}[!ht]
    \centering
    \includegraphics[width=0.8\textwidth]{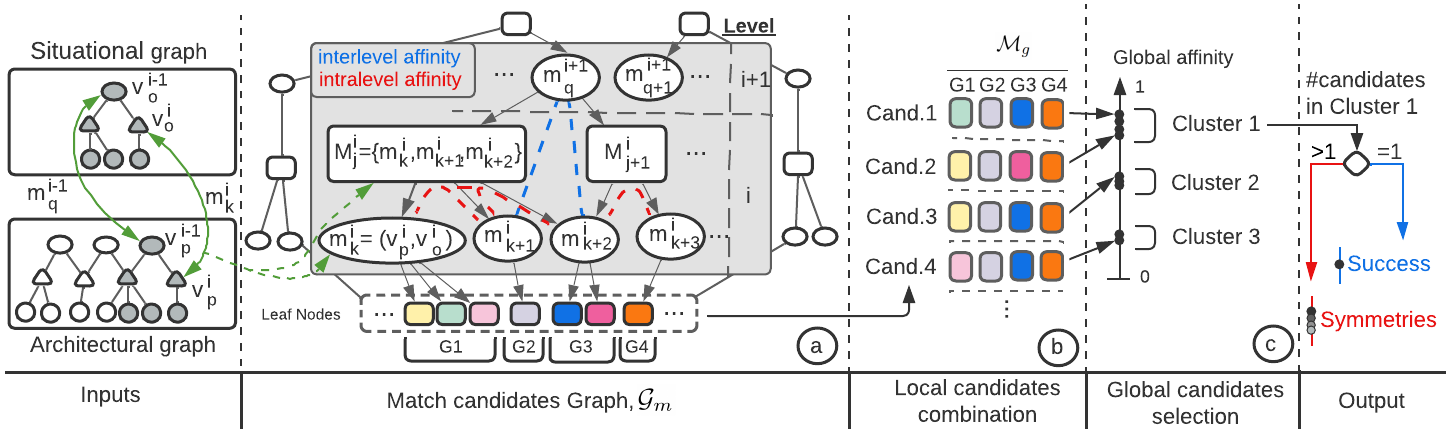}
    \caption{Graph matching schema. a) Downwards and at each level, different combinations of matches are proposed and selected by their geometrical affinity either at the same level or with the associated upper-level pair. b) The match graph is traversed upwards while combined with same-level nodes to define all-level match candidates. c) The lowest-level pairs of every candidate are scored in global affinity. That score is clustered to find symmetries in the best cluster.}
    \label{fig:graph_match_schema}
\end{figure*}

Our second contribution to this paper is a novel approach to graph matching in which we match an architectural \textit{A-graph}, $\mathcal{G}_a$, and a \textit{S-Graph}, $\mathcal{G}_s$. We compare and match the room ($\rho$) entities and for each room entity its corresponding wall-surfaces ($\pi$), extracted from both graphs. The correspondences form a bipartite graph connecting the nodes of some parts of $\mathcal{G}_a$ with all, or almost all nodes, in $\mathcal{G}_s$, at the rooms and walls layers. $\mathcal{G}_s$ is built incrementally as the robot navigates the environment, the graph matching is run after every map update until a successful match is obtained. 
A schema of the entire graph-matching process is described in Fig.~\ref{fig:graph_match_schema}.

\textbf{Notation.} Let $\boldsymbol{V}$ be any set of nodes, with $\boldsymbol{V}_a$ and $\boldsymbol{V}_s$ the sets of nodes in the \textit{A-Graph} and the \textit{S-Graph} respectively. Let $\boldsymbol{m} = {(v_a, v_s) : v_a \in V_a, v_s \in V_s}$, and $\boldsymbol{M}$ be any set of $m$ such as $\boldsymbol{M} =  \{\boldsymbol{m}_1, \boldsymbol{m}_2, ..., \boldsymbol{m}_n\}$. Let $\boldsymbol{\mathcal{M}}$ be any set of  $\boldsymbol{M}$ such as $\boldsymbol{\mathcal{M}} = \{\boldsymbol{M}_1, \boldsymbol{M}_2, ..., \boldsymbol{M}_n\}$. Local candidates are $\boldsymbol{M}$ including $\boldsymbol{m}$ referring to a small part of the input graphs. Global candidates are $\boldsymbol{M}$ including $\boldsymbol{m}$ referring to the entire input graphs. 

\section{Graph Merging}
\label{sec:graph_merging}

Our global state $\mathbf{s}$ at time $T$, before graph merging, contains all the nodes of the \textit{A-Graph}, generated offline, as well as the current nodes estimated online by the \textit{S-Graph}

\begin{eqnarray}
\mathbf{s} &=& [\leftidx{^M}{\mathbf{x}}_{R_1}, \ \hdots, \ \leftidx{^M}{\mathbf{x}}_{R_T}, \nonumber \\
 & & \leftidx{^M}{\boldsymbol{\pi}}_{1}, \ \hdots, \ \leftidx{^M}{\boldsymbol{\pi}}_{P}, \ \leftidx{^B}{\boldsymbol{\pi}}_{1}, \ \hdots, \ \leftidx{^B}{\boldsymbol{\pi}}_{Q}, \nonumber \\
& & \leftidx{^M}{\boldsymbol{\rho}}_{1}, \ \hdots, \ \leftidx{^M}{\boldsymbol{\rho}}_{S}, \ \leftidx{^B}{\boldsymbol{\rho}}_{1}, \ \hdots, \ \leftidx{^B}{\boldsymbol{\rho}}_{R} \\
& & \leftidx{^M}{\boldsymbol{\gamma}}_{1}, \ \hdots, \  \leftidx{^M}{\boldsymbol{\gamma}}_{G},   
\leftidx{^M}{\boldsymbol{\xi}}_{1}, \ \hdots, \ \leftidx{^M}{\boldsymbol{\xi}}_{E}, \nonumber \\
& & \leftidx{^B}{\boldsymbol{\omega}_{1}}, \ \hdots, \  \leftidx{^B}{\boldsymbol{\omega}_{W}}, \  \leftidx{^B}{\boldsymbol{\mathcal{D}}}_{1}, \ \hdots, \  \leftidx{^B}{\boldsymbol{\mathcal{D}}}_{D}, \nonumber
\\
& & \leftidx{^B}{\mathbf{x}}_{M}]^\top, \nonumber
\end{eqnarray}

\noindent where $\leftidx{^B}{\mathbf{x}}_{M}$ is the estimated transformation between the map frame $M$ of the \textit{S-Graph} and the BIM frame \textit{B} of the \textit{A-Graph}, which is set to identity before graph merging. 
The graph matching method from Section.~\ref{sec:graph_matching} provides match candidates between the room nodes and the wall-surface nodes of the \textit{S-Graph} and the \textit{A-Graph}. To efficiently merge the two graphs to generate the \textit{iS-Graph}, we introduce room-to-room constraints as well as a wall-surface-to-wall-surface constraints between the matched candidates. The room-to-room constraint is defined as

\begin{equation}
  c_{\boldsymbol{\rho}} (\leftidx{^B}{\boldsymbol{\rho}_{1}},\leftidx{^M}{\boldsymbol{\rho}_{2}}) = \| \ \leftidx{^B}{\hat{\boldsymbol{\rho}}_{{1}}} - \leftidx{^M}{\hat{\boldsymbol{\rho}}_{{2}}} \|^2_{\mathbf{\Lambda}_{\boldsymbol{\tilde{\boldsymbol{\rho}}}_{1,2}}}~,
\end{equation} 

\noindent where $\leftidx{^B}{\boldsymbol{\rho}_{1}}$ is the room node in the \textit{A-Graph} and $\leftidx{^M}{\boldsymbol{\rho}_{2}}$ is the corresponding room node in \textit{S-Graph}. Similarly, for all correspondences between wall-surface candidates, the constraint is formulated as

\begin{equation}
  c_{\boldsymbol{\pi}} (\leftidx{^B}{\boldsymbol{\pi}_{1}},\leftidx{^M}{\boldsymbol{\pi}_{2}}) = \| \ \leftidx{^B}{\boldsymbol{\pi}_{1}} - \leftidx{^M}{\boldsymbol{\pi}_{2}} \|^2_{\mathbf{\Lambda}_{\boldsymbol{\tilde{\boldsymbol{\pi}}}_{1,2}}}~,
\end{equation}

\noindent where $\leftidx{^B}{\boldsymbol{\pi}_{1}}$ and $\leftidx{^M}{\boldsymbol{\pi}_{2}}$ are the wall-surfaces in the \textit{A-Graph} and the \textit{S-Graph}. 

With the constraints between the two graphs, $\leftidx{^B}{\mathbf{x}}_{M}$ can be estimated and all the robot poses, the wall-surfaces, rooms, and floors of the \emph{S-Graph} can be referred accurately with respect to the BIM frame of reference $B$ of the \emph{A-Graph}, resulting in the final improved situational graph \textit{iS-Graph}. In this manner the robot is localized with respect to the global reference of the architectural plan. Our approach thus can perform global localization exploiting the hierarchical high-level information in the environment without the need for appearance-based loop closure constraints at keyframe level, which are more variable.

\section{Experimental Evaluation}
\label{experimental_evaluation}
\subsection{Experimental Setup}
We validated our proposed approach in multiple simulated and real-world construction environments. 
In all experiments, we generated \textit{A-Graphs} from various building models that were created in Autodesk Revit. The Boston Dynamics \textit{Spot} robot, equipped with a Velodyne VLP-16 3D LiDAR, was teleoperated for data collection. 
We compared our approach to two 2D LiDAR-based localization algorithms (AMCL \cite{AMCL} and Cartographer \cite{cartographer}) and one 3D LiDAR-based localization (UKFL \cite{hdl_graph_slam}) algorithm. In simulated datasets we measured the Absolute Pose Error (APE) \cite{evo_traj_calc} with respect to the ground truth, while in real-world due to the absence of ground truth pose information, we compare the point cloud RMSE of the generated map with the available ground truth map from BIM. 
The proposed methodology was implemented in C++, and the experiments were validated on an Intel i9 16-core workstation.

\begin{table}[b]
\centering
\caption{Absolute Pose Error (APE) [m] for  several LiDAR-based localization baselines and our \textit{iS-Graphs} Datasets have been recorded in simulated environments. `$-$' stands for localization failure.}
\begin{tabular}{c|c c c c c c}
\toprule
\textbf{Method} &  & \multicolumn{4}{c}{\textbf{APE [m] $\downarrow$}}   \\
\toprule
 & & \multicolumn{4}{c}{\textbf{Datasets}} \\ 
 \midrule
  & D1 & D2 & D3 & D4 & D5 & D6   \\ 
\midrule
AMCL \cite{AMCL} & 2.04 & 1.71 & 2.03 & 2.01 & - & -   \\ 
UKFL \cite{hdl_graph_slam} &  0.97 & 0.78 & - & 0.74 & 0.70 & 0.88 \\
Cartographer \cite{cartographer} &   0.10 & 0.16 & 0.12 & - & 0.13 & -\\
\textit{iS-Graphs (ours)} &  \textbf{0.08} & \textbf{0.01} & \textbf{0.02} & \textbf{0.04} &   \textbf{0.09} &   \textbf{0.12} \\
\bottomrule
\end{tabular}
\label{tab:ape_simulation}
\end{table}

\begin{table}[t] 
\setlength{\tabcolsep}{4pt}
\caption{Point cloud RMSE [m] for our real-world dataset. Best results are  boldfaced. `$-$' stands for localization failure.}
\small
\centering
\begin{tabular}{l | c c c c c}
\toprule 
\textbf{Method} & \multicolumn{3}{l}{\textbf{Alignment Error $\downarrow$}} & & \\
\toprule
  &  & \textbf{ Datasets}  &    \\ 
\midrule
  & {D1} &  {D2}  & {D3}   \\ 
\midrule
AMCL \cite{AMCL}  &  - & 0.90 &  0.98  \\ 
UKFL \cite{hdl_graph_slam}  & - & 0.86 & 0.69 \\ 
Cartographer \cite{cartographer} & - & 0.58 & 0.64  \\
\textit{iS-Graphs (ours)} & \textbf{0.17} & \textbf{0.20} & \textbf{0.21}   \\ 
\bottomrule 
\end{tabular}
\label{tab:pcerror_real_data}
\end{table}
\label{ex_grapgh_gen}


\subsection{Results and Discussion}

\textbf{Simulated Datasets.}
Table.~\ref{tab:ape_simulation} presents the APE of our proposed \textit{iS-Graphs} and state-of-the LiDAR-based localization algorithms for six simulated datasets. Each simulated dataset represents a single floor level of a given construction environment with varying configurations of wall-surfaces and rooms. Given the output from 2D LiDAR algorithms, the APE of all the algorithms is computed in 2D ($x$, $y$, $\theta$). Table.~\ref{tab:ape_simulation} shows that our \textit{iS-Graphs} shows higher robustness against localization failure and outperforms all the localization baselines using both 2D and 3D LiDARs.

\textbf{Real-World Datasets.}
Table.~\ref{tab:pcerror_real_data} presents the point cloud RMSE for three different construction environments, where our \textit{iS-Graphs} is able to localize the robot correctly while providing a more accurate 3D map of the environment as compared to the ground truth. Because of noisy LiDAR measurements and the clutter in a real construction environment traditional approaches to localization fail to localize the robot (see D1 in Table.~\ref{tab:pcerror_real_data}).
Since our approach relies only on high-level entities like wall-surfaces and rooms and their topological relationship instead of directly relying on low-level LiDAR measurements, it is more robust to noise and clutter in the environment than the baselines.
\section{Conclusion}
\label{conclusion}
In this paper, we presented a novel method for global robot localization utilizing prior information from architectural plans. We embeded the architectural data from BIM models into optimizable graphs called Architectural Graphs (\textit{A-Graphs}). Using Situational Graphs (\textit{S-Graphs}) estimated by a robot as it navigates its environment, we present a novel graph matching strategy to match the \textit{A-Graphs} with the \textit{S-Graphs} and we also present a graph merging strategy to fuse both graphs. The result of the fusion is an informed \textit{iS-Graph}, which enables the robot to localize itself within the architectural plan for the given environment.

\balance
\bibliographystyle{IEEEtran}
\bibliography{root}

\end{document}